\documentclass[sigconf,nonacm]{acmart}

\usepackage{booktabs}
\usepackage{multirow}
\usepackage{algorithm}
\usepackage{algpseudocode}
\usepackage{subcaption}
\usepackage{makecell}
\usepackage{graphicx}
\usepackage{amsmath}

\usepackage{amssymb}
\usepackage[colorinlistoftodos]{todonotes}
\usepackage{pifont}
\newcommand{\cmark}{\ding{51}}
\newcommand{\xmark}{\ding{55}}

\setcopyright{acmlicensed}
\copyrightyear{2026}
\acmYear{2026}

\title{Transferable Latency Prediction for Fast LLM Screening on Heterogeneous Edge Devices}

\author{Xiaolong Tu}
\affiliation{%
  \institution{Georgia State University}
  \city{Atlanta}
  \state{GA}
  \country{USA}
}
\email{xtu@gsu.edu}

\author{Vinod K. Mishra}
\affiliation{%
  \institution{DEVCOM Army Research Laboratory}
  \city{Aberdeen Proving Ground}
   \state{MD}
  \country{USA}
}
\email{vinod.k.mishra.civ@army.mil}

\author{Venkat R. Dasari}
\affiliation{%
  \institution{DEVCOM Army Research Laboratory}
   \city{Aberdeen Proving Ground}
   \state{MD}
  \country{USA}
}
\email{venkateswara.r.dasari.civ@army.mil}

\author{Anu G. Bourgeois}
\affiliation{%
  \institution{Georgia State University}
  \city{Atlanta}
  \state{GA}
  \country{USA}
}
\email{abourgeois@gsu.edu}

\author{Haoxin Wang}
\affiliation{%
  \institution{Georgia State University}
  \city{Atlanta}
  \state{GA}
  \country{USA}
}
\email{hwang@gsu.edu}

\begin{abstract}
Accurate latency prediction is critical for deploying large language models (LLMs) on heterogeneous edge devices, where inference latency is affected by model architecture, prompt behavior, runtime backend, hardware utilization, dynamic voltage and frequency scaling (DVFS), and thermal variation. This paper presents a runtime-aware latency prediction framework for deployment-oriented LLM selection. The framework represents each inference request as a hardware-runtime-model-prompt configuration, separates inference into prefill and decode phases, and adaptively fuses static descriptors with dynamic hardware telemetry through a gated prediction model. 

We evaluate the framework using Pixel mobile devices and validate the profiling pipeline on Jetson Nano, Orange Pi 5 Pro, and an RTX~3090-class GPU platform. On Pixel~8, the full predictor improves total-latency $R^2$ from 0.953 to 0.960 and decode-latency $R^2$ from 0.957 to 0.973 over a static-only baseline. On Pixel~8~Pro, it improves prefill-latency $R^2$ from -1.383 to 0.966. For cross-device transfer, calibration improves Pixel~8~Pro $\rightarrow$ Pixel~8 total-latency $R^2$ from -0.974 to 0.940 and decode-latency $R^2$ from -1.085 to 0.927. Heterogeneous profiling further shows that latency is highly device- and runtime-dependent: the same SmolLM2 model family reaches 8.42 tokens/s on Orange Pi 5 Pro but 64.38 tokens/s on an RTX~3090-class GPU. 
These results demonstrate that runtime-aware prediction with lightweight calibration can reduce profiling cost and support latency-aware LLM deployment across heterogeneous edge platforms.
\end{abstract}


\renewcommand\footnotetextcopyrightpermission[1]{}
\begin{document}
\maketitle

\section{Introduction}

Large language models (LLMs) are increasingly attractive for deployment on edge and on-device platforms to support low-latency, privacy-sensitive, and connectivity-constrained applications~\cite{li2024mobilellm,xu2024ondevice}. In mission-critical and tactical environments, local inference is especially important because cloud offloading may be limited by intermittent connectivity, bandwidth constraints, security concerns, or strict response deadlines. Recent inference systems and profiling tools, such as MLC-LLM, lm-Meter, and local PyTorch/Hugging Face backends, have made it increasingly practical to execute and analyze compact LLMs across mobile and edge platforms~\cite{mlcllm2024github,wang2025lmmeter}. However, an important challenge remains unresolved: \emph{how to quickly determine which model is suitable for which device under which deployment requirement}.

This challenge is becoming more significant as modern model repositories provide rapidly growing collections of candidate LLMs with diverse architectures, sizes, quantization formats, and runtime characteristics. Although this ecosystem offers deployment flexibility, it also creates a major systems bottleneck. A model's latency depends not only on its architecture, but also on the target device, prompt workload, runtime backend, execution platform, and current hardware state~\cite{xiao2024pockets}. Therefore, a model that satisfies a latency budget on one device-runtime environment may fail to meet the same requirement on another.
A straightforward solution is to deploy and benchmark every candidate model on every target device. However, exhaustive profiling is expensive because it requires repeated runtime setup, model conversion, device-side execution, telemetry collection, and post-processing~\cite{zhang2021nnmeter,tu2023energy,tu2025aienergy}. This cost grows quickly with the number of candidate models, devices, prompts, and runtime configurations. A practical deployment system therefore needs a latency prediction capability that can screen candidate models before full target-device execution.
Existing latency estimation approaches are insufficient for this setting. Predictors based only on static architecture or hardware features can miss platform-specific behavior, runtime scheduling effects, and backend-dependent execution differences~\cite{akhauri2024latency}. For LLM inference, latency is shaped by both \emph{static factors}, such as model architecture, parameter scale, quantization format, and hardware capability, and \emph{dynamic factors}, such as prompt length, generated-token behavior, DVFS, utilization variation, memory pressure, and thermal effects~\cite{kwon2023pagedattention,kim2021ztt,lin2023geardvfs,benoitcattin2020thermal}. In addition, LLM inference has two distinct phases: prefill, which processes the prompt and constructs the key-value cache, and decode, which generates tokens autoregressively. These phases exhibit different latency and hardware behavior, making phase-aware modeling necessary.

In this paper, we present a runtime-aware and transferable latency prediction framework for deployment-oriented LLM selection across heterogeneous device-runtime environments. The framework represents each inference request as a hardware-runtime-model-prompt configuration. It combines static descriptors of devices, models, and deployment settings with dynamic runtime telemetry collected during execution. The predictor separates prefill and decode behavior through phase-aware modeling and adaptively fuses static and dynamic embeddings through a gated fusion module. A lightweight calibration set is then used to align the predictor with target-device latency behavior.
The framework connects latency prediction to deployment-time model screening. Predicted latency is used to filter and rank candidate models before expensive target-device profiling. In the final selection stage, latency can be combined with model accuracy or task quality to identify Pareto-optimal candidates that balance responsiveness and output quality.
We implement and evaluate the framework across Pixel mobile devices, Jetson Nano, Orange Pi 5 Pro, and an RTX~3090-class GPU platform. The mobile dataset provides the main predictor evaluation, while the Jetson, Orange Pi, and RTX~3090 datasets validate that the profiling schema extends to embedded GPU, single-board edge, and high-performance GPU reference platforms. 

\textbf{The main contributions of this paper are as follows:}
\begin{itemize}
    \item We formulate LLM latency prediction as a hardware-runtime-model-prompt configuration problem, enabling prediction across heterogeneous devices, models, prompts, and runtime backends.

    \item We propose a runtime-aware transferable latency predictor that combines static descriptors with dynamic telemetry, separates prefill and decode behavior through phase-aware modeling, and uses gated fusion to combine static and dynamic embeddings.

    \item We design a deployment-oriented screening workflow that uses calibrated latency prediction to filter candidate models before expensive target-device execution and supports latency-quality trade-off analysis for final selection.

    \item We build and evaluate a multi-platform profiling pipeline across Pixel mobile devices, Jetson Nano, Orange Pi 5 Pro, and an RTX~3090-class GPU platform, validating both mobile latency prediction and heterogeneous profiling coverage.
\end{itemize}

\section{Background and Motivation}

\subsection{On-device LLM Inference on Edge Devices}

Running LLMs on mobile and edge devices is attractive because it enables local processing under strict latency, privacy, and connectivity constraints. At the same time, edge deployment remains fundamentally more challenging than server-side deployment because mobile platforms are constrained by limited memory capacity, lower sustained compute throughput, and tighter thermal envelopes. These constraints make latency a first-order concern when selecting a model for a target device and application.

Modern autoregressive LLM inference typically consists of two execution phases: \emph{prefill} and \emph{decode}. In the prefill phase, the model processes the input prompt and constructs the key--value (KV) cache. In the decode phase, the model generates output tokens iteratively using the previously computed KV states. Prior work on LLM serving has emphasized that these two phases exhibit substantially different execution behavior and resource demands, and that separating or scheduling them differently can be important for performance~\cite{yu2022orca,zhong2024distserve}. As a result, the latency of on-device LLM inference cannot be adequately characterized by a single static cost estimate.

\subsection{Hardware Dynamics on Edge Devices}

Unlike server-class platforms, edge devices operate under tight power and thermal budgets. Early measurement studies on smartphones already showed that mobile systems are highly constrained by subsystem-level power behavior and limited energy headroom~\cite{carroll2010smartphone}. Subsequent work on mobile thermal management further demonstrated that aggressive use of available thermal budget can lead to substantially degraded sustained performance over time~\cite{sahin2015greedythermal}. These effects are particularly important for on-device LLM inference, where long prompts, long decoding sequences, and repeated interactive requests can expose significant variation in effective throughput. Consequently, the latency of a candidate LLM on an edge device is determined not only by static model and hardware properties, but also by dynamic runtime conditions such as thermal state, power-management behavior, and workload phase behavior. Recent edge-AI measurement and benchmark studies further show that energy and performance behavior can vary substantially across devices, workloads, and runtime settings, motivating systematic profiling datasets and benchmark suites for mobile and IoT platforms~\cite{tu2023energy,tu2023deepen,tu2025aienergy}.

\subsection{Limitations of Static Latency Predictors}

Latency prediction has long been studied as a key enabler for efficient model deployment on edge hardware. Recent work on hardware-adaptive latency prediction shows that cross-device generalization is difficult when predictors rely too heavily on platform-specific calibration, and that few-shot adaptation or meta-learning can substantially reduce the cost of moving to unseen hardware~\cite{lee2021help}. Likewise, benchmark efforts such as HW-NAS-Bench highlight the diversity of latency behavior across hardware platforms and make clear that hardware-aware model selection cannot rely on architecture information alone~\cite{li2021hwnasbench}. 

For on-device LLM deployment, the limitations of static predictors become even more pronounced. First, LLM latency depends not only on model structure but also on workload configuration, such as prompt length and output length. Second, the latency contributions of prefill and decode differ substantially across deployment scenarios. Third, edge devices introduce dynamic performance variation that is not captured by static descriptors alone. Therefore, a practical predictor for on-device LLM model screening must be both \emph{runtime-aware} and \emph{transferable}: runtime-aware to capture dynamic execution conditions, and transferable to generalize across heterogeneous devices without exhaustive profiling on every new platform.

\section{System Design}
\label{sec:system}

\begin{figure*}[t]
  \centering
  \includegraphics[width=\textwidth]{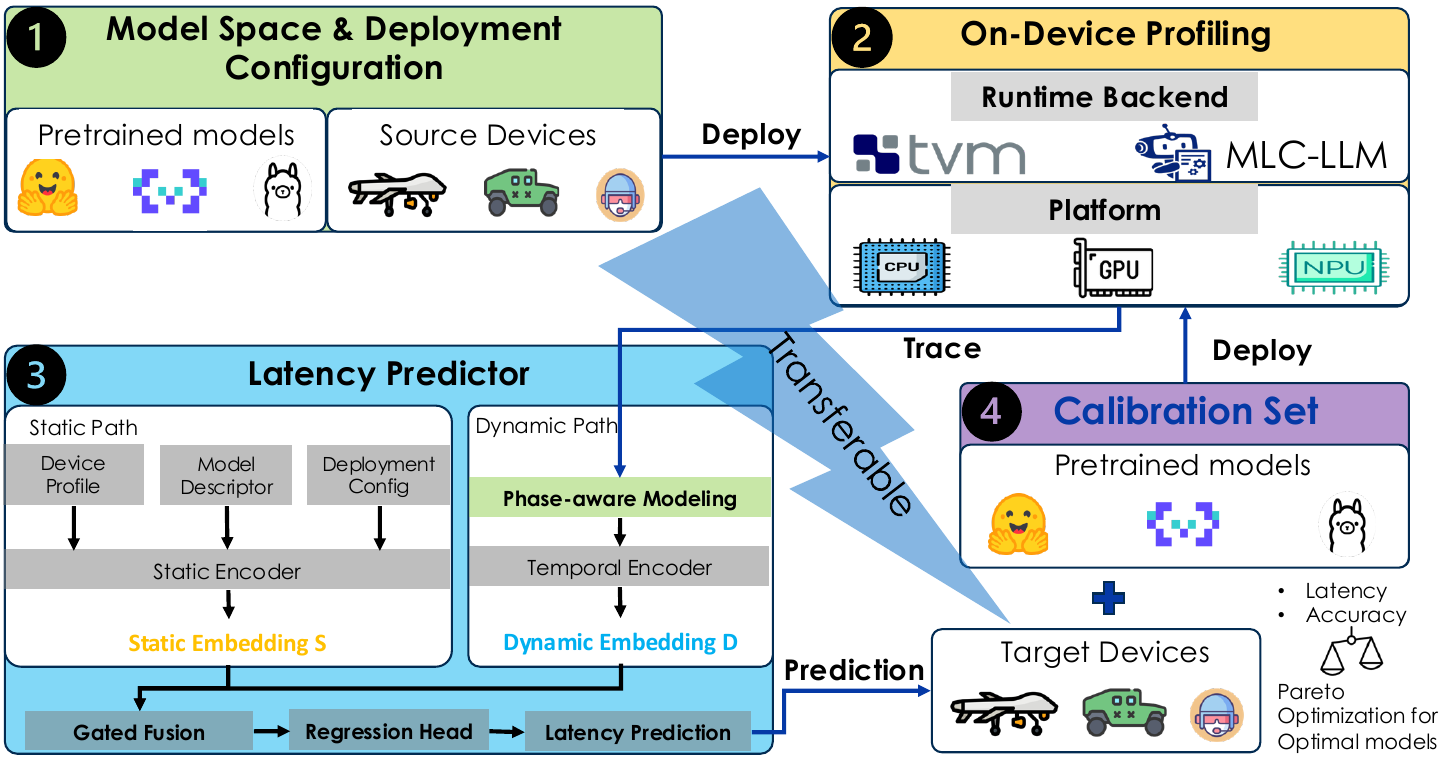}
  \caption{\textbf{Overall system framework.} The system begins with a model space and deployment configuration that includes pretrained models, source devices, target devices, runtime backend, and platform settings. Source-device executions are deployed and profiled to collect device profiles, model descriptors, deployment configurations, and runtime traces. These inputs are processed by a transferable latency predictor with a static path and a dynamic path.}
  \label{fig:system_overview}
\end{figure*}

Figure~\ref{fig:system_overview} presents the overall framework of the proposed system. The system is designed to support latency-aware LLM deployment across heterogeneous devices without exhaustively benchmarking every candidate model on every target platform. It starts from a model space and deployment configuration, profiles representative executions on source devices, trains a transferable latency predictor using both static descriptors and runtime traces, and then applies the predictor to target devices. During deployment, a small calibration set is used to adapt the predictor to the target-device runtime regime. Finally, predicted latency is combined with model quality metrics to perform Pareto-based model selection.

\subsection{Overview}

The goal of the proposed framework is to reduce the cost of deploying LLMs on heterogeneous devices by replacing exhaustive profiling with transferable latency prediction. Given a pool of pretrained models and a set of possible deployment configurations, a naive approach would benchmark every model on every device under every runtime setting. This is expensive because LLM latency depends on multiple interacting factors, including model architecture, prompt length, runtime backend, hardware state, execution platform, and device-specific scheduling behavior.

The proposed framework addresses this problem by learning a latency predictor from previously profiled source-device executions. The predictor maps a hardware-runtime-model-prompt configuration to latency targets, including prefill latency, decode latency, and total request latency. During deployment, the trained predictor estimates latency for candidate models on a target device. A small calibration set collected from the target device is used to correct device-specific bias. The calibrated latency prediction is then combined with model accuracy to identify Pareto-optimal candidate models that balance latency and quality.
The system therefore consists of four major stages: (1) defining the model space and deployment configuration, (2) collecting source-device profiling data, (3) training a transferable latency predictor, and (4) performing deployment-time prediction, calibration, and Pareto-based model selection.

\subsection{Model Space and Deployment Configuration}

The system operates over a structured deployment space composed of pretrained models, source devices, target devices, runtime backends, and platform settings. The model space contains candidate LLMs drawn from model zoos or local model repositories. These models may differ in architecture family, parameter scale, quantization format, and deployment compatibility. The device space contains both \emph{source devices}, which provide profiling data for training, and \emph{target devices}, on which final deployment decisions are made.

Each candidate deployment is represented as a configuration tuple:
\begin{equation}
c = (m, d, r, p),
\end{equation}
where $m$ denotes the pretrained model, $d$ denotes the device, $r$ denotes the runtime backend and platform setting, and $p$ denotes the prompt or task configuration. The runtime backend may include MLC-LLM, PyTorch, Hugging Face Transformers, or other device-specific inference stacks. The platform setting specifies whether inference is executed on CPU, GPU, NPU, or another accelerator when available.
This formulation supports cross-device, cross-model, and cross-prompt prediction. Instead of requiring the same prompt-model pair to be measured on every device, the system treats each inference request as one observed deployment configuration. This is important for practical deployment because profiling data may be collected from different devices, model families, prompt sets, and runtime environments over time.

\subsection{Source-Device On-Device Profiling}

The first data collection stage performs on-device profiling on source devices. For each selected deployment configuration, the system deploys the candidate model through the corresponding runtime backend and records both request-level latency and runtime telemetry. The profiling process produces four classes of information: device profile, model descriptor, deployment configuration, and runtime trace.
The \emph{device profile} describes hardware and platform characteristics, such as device type, processor or accelerator information, memory capacity, frequency range, and available telemetry signals. The \emph{model descriptor} summarizes model-side properties, including model family, parameter scale, architecture configuration, quantization format, and deployment format. The \emph{deployment configuration} captures runtime-level choices, such as backend, platform, maximum generation length, and execution mode. The \emph{runtime trace} captures execution-dependent behavior, including latency, utilization, frequency, temperature, memory usage, power-related signals, and other available hardware measurements.
These profiling records provide the supervision required to train the latency predictor. They also expose the key motivation for runtime-aware prediction: two deployments with similar static model sizes may have different latency because runtime scheduling, hardware utilization, memory behavior, and phase-specific execution patterns can differ substantially across devices.

\subsection{Hardware-Embedding-Based Runtime-Aware Latency Predictor}

The core of the framework is a transferable latency predictor with two complementary branches: a static path and a dynamic path. The static path captures stable deployment properties, while the dynamic path captures runtime-dependent behavior from profiling traces. The two paths are fused by a gated fusion module before final latency prediction.

\subsubsection{Static Path and Static Encoder}

The static path processes features that are known before execution or remain relatively stable for a deployment configuration. These features include the device profile, model descriptor, and deployment configuration. Examples include device type, accelerator type, memory capacity, model family, parameter scale, quantization format, runtime backend, platform setting, input length, and maximum generation length.

Let $\mathbf{x}_s$ denote the static feature vector. The static encoder maps this vector into a compact static embedding:
\begin{equation}
\mathbf{S} = f_s(\mathbf{x}_s),
\end{equation}
where $f_s(\cdot)$ is the static encoder and $\mathbf{S}$ is the static embedding. This embedding represents the expected compatibility between the model, device, and runtime configuration before considering detailed execution-time behavior.

\subsubsection{Dynamic Path, Phase-Aware Modeling, and Temporal Encoder}

The dynamic path models execution-dependent behavior using runtime traces collected during profiling. LLM inference has two structurally different phases: \emph{prefill} and \emph{decode}. Prefill processes the input prompt and constructs the key-value cache, while decode generates tokens autoregressively. These two phases have different computational and memory-access patterns. Prefill is often shorter and more bursty, whereas decode is longer, sequential, and more sensitive to token-by-token runtime behavior.

\begin{figure*}[t]
  \centering
  \includegraphics[width=\textwidth]{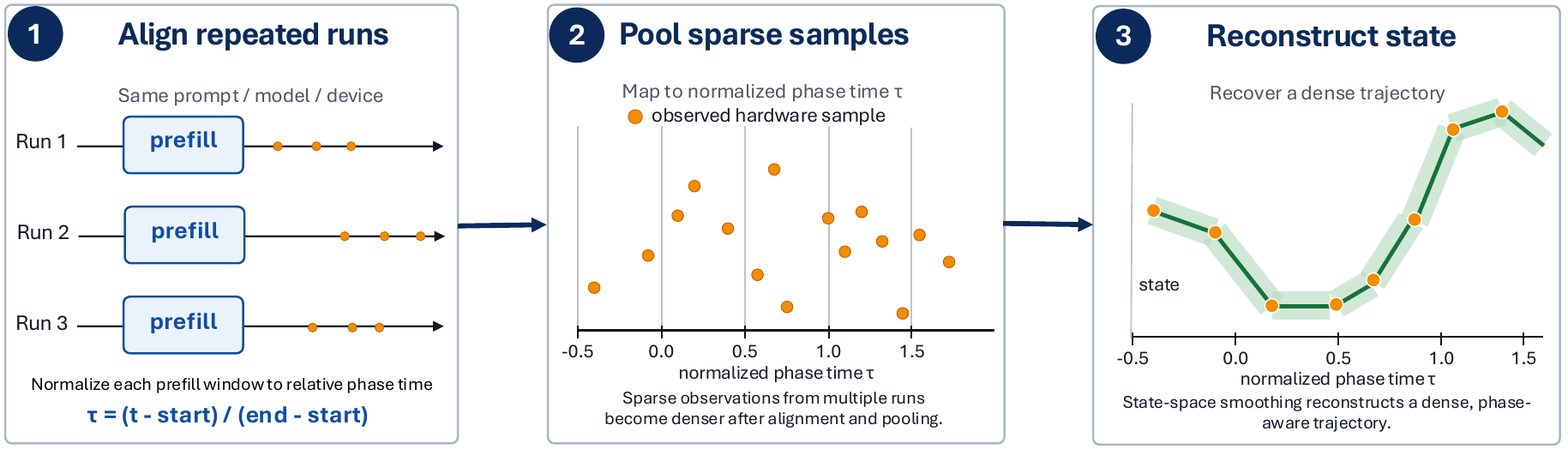}
  \caption{\textbf{Phase-aware modeling of runtime traces.} The raw runtime trace is reorganized into phase-structured representations for prefill and decode. Prefill traces from repeated runs are aligned under normalized phase time to reconstruct sparse hardware states, while decode traces preserve autoregressive temporal behavior for sequential modeling.}
  \label{fig:phase_aware_modeling}
\end{figure*}

To preserve this structure, the system applies phase-aware modeling before temporal encoding. The raw trace is segmented into prefill and decode regions. Prefill observations are aligned into a phase-level representation to reduce noise caused by sparse and bursty measurements. Decode observations preserve temporal order so that the model can capture autoregressive execution behavior.
After phase-aware preprocessing, the processed runtime sequence is passed to the temporal encoder. The temporal encoder projects per-step runtime observations into latent states, captures temporal dependencies, and aggregates them into a dynamic embedding. Let $\mathbf{x}_d$ denote the processed runtime sequence. The temporal encoder produces:
\begin{equation}
\mathbf{D} = f_d(\mathbf{x}_d),
\end{equation}
where $f_d(\cdot)$ is the temporal encoder and $\mathbf{D}$ is the dynamic embedding. The dynamic embedding captures hardware state variation, utilization behavior, thermal effects, memory activity, and phase-specific runtime patterns relevant to latency prediction.

\subsubsection{Gated Fusion and Regression Head}

The static and dynamic paths provide complementary information. Static descriptors are useful for estimating latency before execution, while dynamic traces explain runtime variation that cannot be inferred from static features alone. The predictor combines the static embedding $\mathbf{S}$ and dynamic embedding $\mathbf{D}$ through a gated fusion module.

The gated fusion module computes adaptive weights:
\begin{equation}
[w_s, w_d] = \mathrm{softmax}(g([\mathbf{S};\mathbf{D}])),
\end{equation}
where $g(\cdot)$ is a lightweight gating network and $[\mathbf{S};\mathbf{D}]$ denotes concatenation. The fused representation is:
\begin{equation}
\mathbf{F} = w_s \mathbf{S} + w_d \mathbf{D}.
\end{equation}

This adaptive fusion allows the predictor to rely more on static information when the deployment configuration is stable and more on dynamic information when runtime behavior dominates latency variation. The fused representation is then passed to a regression head:
\begin{equation}
\hat{\mathbf{y}} = h(\mathbf{F}),
\end{equation}
where $\hat{\mathbf{y}}$ contains the predicted prefill latency, decode latency, and total latency. The regression head can be implemented as a multi-output predictor so that phase-level and total latency targets are learned jointly.

\subsection{Transferable Latency Predictor Training}

The predictor is trained using profiling records collected from source devices. Each training sample contains static descriptors, phase-aware runtime features, and measured latency targets. The training objective encourages the predictor to learn representations that are useful across devices, models, prompts, and runtime backends.

Let $\mathbf{y}$ denote the measured latency target and $\hat{\mathbf{y}}$ denote the predicted latency. We optimize a weighted combination of mean absolute error and mean squared error:
\begin{equation}
\mathcal{L}_{\text{lat}} =
\alpha \cdot \mathrm{MAE}(\mathbf{y},\hat{\mathbf{y}})
+
(1-\alpha) \cdot \mathrm{MSE}(\mathbf{y},\hat{\mathbf{y}}),
\end{equation}
where $\alpha \in [0,1]$ balances robustness to outliers and sensitivity to large latency errors.

Because prefill and decode have different execution patterns, the framework can further apply phase-level supervision:
\begin{equation}
\mathcal{L}_{\text{phase}} =
\lambda_{\text{pre}}\mathcal{L}_{\text{pre}}
+
\lambda_{\text{dec}}\mathcal{L}_{\text{dec}},
\end{equation}
where $\mathcal{L}_{\text{pre}}$ and $\mathcal{L}_{\text{dec}}$ correspond to prefill and decode latency losses. The final objective is:
\begin{equation}
\mathcal{L} =
\mathcal{L}_{\text{lat}}
+
\mathcal{L}_{\text{phase}}.
\end{equation}

This training formulation helps the predictor learn both total request-level behavior and phase-specific latency patterns. It is especially important for transfer because the relative contribution of prefill and decode may shift across devices and runtime backends.

\subsection{Target-Device Prediction and Calibration}

After training, the predictor is applied to candidate deployments on a target device. For each target-device candidate, the system constructs the static input from the device profile, model descriptor, and deployment configuration. If a small calibration set is available, the system also collects limited target-device traces to estimate dynamic behavior and correct target-specific latency bias.

The calibration set contains a small number of measured target-device executions. These samples are not intended to replace full profiling. Instead, they serve as anchors that align the source-trained predictor with the latency scale and runtime behavior of the target device. Calibration can be implemented by fine-tuning lightweight layers, fitting an output correction model, or updating normalization parameters using target-device measurements.
This calibration loop is important because cross-device transfer is affected by device-specific runtime effects. Even when the same model and prompt are used, different hardware schedulers, memory systems, thermal states, and runtime backends can shift latency distributions. The calibration set reduces this mismatch while preserving the main benefit of the system: avoiding exhaustive profiling of every candidate deployment.

\subsection{Pareto-Based Deployment Selection}

The final stage converts latency prediction into a deployment decision. The system does not optimize latency alone because the fastest model may not provide acceptable response quality. Instead, deployment selection is formulated as a multi-objective optimization problem that jointly considers predicted latency and model quality.

Let $\mathcal{M}=\{m_1,m_2,\ldots,m_N\}$ denote the set of candidate models for a target device. For each candidate model $m_i$, the calibrated predictor estimates a latency value $\hat{L}_i$, and an external evaluation procedure provides an accuracy or task-quality score $A_i$. The deployment objective can be written as:
\begin{equation}
\min_{m_i \in \mathcal{M}} \hat{L}_i,
\qquad
\max_{m_i \in \mathcal{M}} A_i.
\end{equation}
These two objectives are generally conflicting: a larger model may produce better responses but incur higher latency, while a smaller model may be faster but less accurate. Therefore, there may not exist a single model that is simultaneously best for both objectives.
To reason about this trade-off, we use Pareto dominance. Candidate $m_i$ dominates candidate $m_j$ if $m_i$ is no worse than $m_j$ in all objectives and strictly better in at least one objective. In our latency-quality setting, this is defined as:
\begin{equation}
\hat{L}_i \leq \hat{L}_j,
\qquad
A_i \geq A_j,
\end{equation}
with at least one strict inequality:
\begin{equation}
\hat{L}_i < \hat{L}_j
\quad \text{or} \quad
A_i > A_j.
\end{equation}
If $m_i$ dominates $m_j$, then $m_j$ is never preferable under any deployment policy that favors lower latency and higher quality, because $m_i$ provides equal or better latency and equal or better quality.

The Pareto frontier is the set of non-dominated candidates:
\begin{equation}
\mathcal{P}
=
\left\{
m_i \in \mathcal{M}
\;\middle|\;
\nexists m_j \in \mathcal{M}
\text{ such that }
m_j \succ m_i
\right\},
\end{equation}
where $m_j \succ m_i$ denotes that $m_j$ dominates $m_i$. Each model on the Pareto frontier represents a valid trade-off between latency and quality. Moving along the frontier improves one objective only by sacrificing the other. Therefore, the Pareto frontier provides a compact set of candidate deployments that are worth considering, while dominated models can be safely filtered out before expensive target-device execution.
This Pareto optimization step allows the system to reduce deployment search cost. Instead of exhaustively profiling all candidate models on the target device, the system first predicts latency, combines it with model quality, and removes candidates that are dominated. The remaining Pareto-optimal or near-Pareto-optimal candidates are then prioritized for target-device validation. In this way, predicted latency is not used as an isolated regression output, but as part of a deployment decision process that balances efficiency and quality under device-specific constraints.

\subsection{Deployment-Time Closed Loop}

The system forms a closed deployment loop. Prediction reduces the candidate search space, profiling selected candidates provides calibration data, and calibration improves future predictions on the same target device. This loop allows the system to progressively improve deployment decisions with limited measurements.
In practice, the target-device workflow proceeds as follows. First, the system predicts latency for candidate models under the target deployment configuration. Second, it identifies promising candidates using Pareto optimization over predicted latency and accuracy. Third, a small number of selected candidates are deployed and profiled on the target device. Fourth, the resulting traces update the calibration set and refine future predictions. Through this iterative process, the system moves from source-device knowledge to target-device deployment decisions while keeping profiling cost low.

\section{Experimental Design}
\label{sec:exp}

We evaluate the proposed framework from three complementary perspectives: latency prediction accuracy, transferability across heterogeneous deployment configurations, and deployment-oriented model selection. The experiments are designed to answer three practical questions: (1) can the predictor estimate prefill, decode, and total latency for LLM inference under different device-runtime conditions, (2) can the learned representation transfer across devices, models, prompts, and runtime backends with limited target-device calibration, and (3) can predicted latency, together with model quality, reduce deployment-time search cost while preserving strong candidate models?

This design follows the system pipeline described in Section~\ref{sec:system}. Source-device profiling provides device profiles, model descriptors, deployment configurations, and runtime traces. The predictor uses these inputs to estimate phase-level and total latency. A small calibration set aligns the predictor with target-device runtime behavior. Finally, predicted latency is used for deployment-oriented screening and can be combined with model quality for Pareto-based model selection.


\subsection{Experimental Setup}

\textbf{Devices.}
The evaluation covers four heterogeneous device-runtime environments: Pixel mobile devices, NVIDIA Jetson Nano, Orange Pi 5 Pro, and an RTX~3090-class GPU platform. These platforms represent complementary deployment classes, including mobile SoCs, embedded GPU devices, single-board edge platforms, and high-performance GPU reference hardware. Pixel mobile devices represent on-device LLM deployment with Android runtime overhead, mobile scheduling behavior, memory bandwidth limits, and thermal variation. Jetson Nano represents low-power GPU-based edge inference. Orange Pi 5 Pro represents lightweight ARM-based single-board inference. The RTX~3090-class platform provides a high-throughput reference regime with substantially different latency, utilization, and power behavior.
The devices play two roles in the evaluation. Pixel~8 and Pixel~8~Pro provide the main prediction and transfer evaluation because they include cleaned request-level samples and complete predictor evaluation results. Jetson Nano, Orange Pi 5 Pro, and the RTX~3090-class platform are used to validate whether the same profiling schema can extend across heterogeneous runtime and hardware environments. This design connects mobile prediction accuracy with broader device-runtime profiling coverage.

\begin{table*}[t]
\centering
\caption{Device setup in the experimental design.}
\label{tab:device_setup}
\footnotesize
\setlength{\tabcolsep}{5pt}
\renewcommand{\arraystretch}{1.12}
\begin{tabular}{p{0.20\textwidth} p{0.23\textwidth} p{0.34\textwidth} p{0.16\textwidth}}
\toprule
Device & Deployment class & Motivation & Evaluation role \\
\midrule
Pixel~8 / Pixel~8~Pro 
& Mobile SoC platform 
& Represents on-device mobile LLM deployment with Android runtime overhead, thermal variation, memory bandwidth limits, and mobile scheduling behavior 
& Prediction and transfer evaluation \\
Jetson Nano 
& Embedded GPU edge platform 
& Represents low-power GPU-based edge inference with different runtime and hardware behavior from mobile devices 
& Profiling validation \\
Orange Pi 5 Pro 
& Single-board edge platform 
& Represents lightweight ARM-based local inference with constrained compute and memory resources 
& Profiling validation \\
RTX~3090-class GPU 
& High-performance GPU reference 
& Provides a high-throughput reference regime for comparing latency, utilization, and runtime behavior against edge platforms 
& Profiling validation \\
\bottomrule
\end{tabular}
\end{table*}

\textbf{Models.}
The model space consists of compact pretrained LLMs spanning different model families, parameter scales, and runtime formats. The reported model pool includes \emph{SmolLM2-135M-Instruct}, \emph{SmolLM2-360M-Instruct}, \emph{SmolLM2-1.7B-Instruct}, \emph{TinyLlama-1.1B-Chat}, and \emph{OPT-350M}. These models cover compact LLM families and parameter scales from approximately 135M to 1.7B parameters.
We organize the evaluation around a model pool rather than a single fixed model because the final objective is deployment-oriented model screening. In a practical deployment scenario, the system must estimate whether a candidate model can satisfy device-specific latency constraints before full target-device profiling. Therefore, the predictor is expected to generalize across model instances and provide latency estimates that are useful for ranking, filtering, and selecting candidate models under hardware-specific constraints.

\begin{table*}[t]
\centering
\caption{Model setup in the experimental design.}
\label{tab:model_setup}
\footnotesize
\setlength{\tabcolsep}{4pt}
\renewcommand{\arraystretch}{1.08}
{
\begin{tabular}{llll}
\toprule
Model & Approx. scale & Model family / source & Role in evaluation \\
\midrule
SmolLM2-135M-Instruct & 135M & SmolLM2 & Small compact LLM for low-resource edge settings \\
SmolLM2-360M-Instruct & 360M & SmolLM2 & Main compact model for mobile and edge experiments \\
OPT-350M & 350M & OPT / Meta & Decoder-only transformer baseline \\
TinyLlama-1.1B-Chat & 1.1B & TinyLlama & Larger compact chat model baseline \\
SmolLM2-1.7B-Instruct & 1.7B & SmolLM2 & Larger compact LLM for scaling analysis \\
\bottomrule
\end{tabular}
}
\end{table*}

\textbf{Inference requests and telemetry.}
Each profiling record corresponds to one LLM inference request under a specific device, runtime, model, and prompt configuration. The prompt set is treated as part of the configuration space rather than as a fixed benchmark that must be identical across all devices. For each request, we record prompt-side descriptors, including prompt identifier, task category when available, input token count, generated token count, maximum generation length, and runtime configuration.
During each inference run, the system records runtime telemetry from the available hardware monitoring interface of each platform. Although the exact telemetry fields differ across devices, they are mapped into a unified runtime-aware representation that captures utilization, frequency, memory, temperature, and power-related behavior at the request level. This allows the predictor to learn from heterogeneous traces without requiring identical hardware counters on every platform.

\textbf{Latency targets.}
We predict three latency targets: prefill latency, decode latency, and total request latency. Prefill latency measures the time required to process the input prompt and construct the key-value cache. Decode latency measures the autoregressive generation phase after prefill. Total request latency measures end-to-end latency. Separating prefill and decode is important because prefill is dominated by prompt processing and parallel sequence computation, whereas decode is dominated by iterative token generation, cache access, and memory behavior.

\subsection{Baselines and Ablation Settings}

We compare the full predictor against progressively stronger baselines and ablated variants to isolate the contribution of different components. The \emph{static-only baseline} uses only static deployment descriptors, including device descriptors, model descriptors, runtime backend, platform setting, prompt-level descriptors, and generation configuration. The \emph{runtime-aware model without phase-aware modeling} includes runtime telemetry but does not explicitly preserve the prefill/decode phase structure. The \emph{fusion without gating} variant includes both static and dynamic information but replaces gated fusion with direct concatenation or a fixed combination of the two branches. The \emph{full model without calibration} uses the complete feature set and predictor structure but directly applies the trained predictor to the target device without target-device refinement. The \emph{full model with calibration} uses the complete feature set together with a small calibration set from the target device before final testing.

\begin{table*}[t]
\centering
\caption{Baseline and ablation settings for configuration-level latency prediction.}
\label{tab:baseline_ablation}
\footnotesize
\setlength{\tabcolsep}{4pt}
\renewcommand{\arraystretch}{1.08}
{
\begin{tabular}{lccccc}
\toprule
Variant & Static descriptors & Runtime-aware features & Phase-aware modeling & Gated fusion & Calibration \\
\midrule
Static-only baseline & \cmark & \xmark & \xmark & \xmark & \xmark \\
Runtime-aware w/o phase-aware modeling & \cmark & \cmark & \xmark & \xmark & \xmark \\
Fusion w/o gating & \cmark & \cmark & \cmark & \xmark & \xmark \\
Full model w/o calibration & \cmark & \cmark & \cmark & \cmark & \xmark \\
Full model w/ calibration & \cmark & \cmark & \cmark & \cmark & \cmark \\
\bottomrule
\end{tabular}
}
\end{table*}

\subsection{Evaluation Protocol and Metrics}

We evaluate the framework under three settings. First, the \emph{within-domain setting} randomly partitions requests from the same device-runtime-model-prompt distribution into training and testing sets, measuring standard latency prediction accuracy. Second, the \emph{transfer setting} holds out one device, model family, prompt group, or heterogeneous configuration as the target domain and trains on the remaining data. This setting measures cross-device, cross-model, and cross-prompt generalization. When target-domain samples are available, we evaluate both zero-shot prediction and calibrated prediction using a small target-domain calibration set. Third, the \emph{deployment screening setting} uses predicted latency to reduce the candidate model space before real target-device execution.
For latency prediction, we report mean absolute error (MAE), root mean squared error (RMSE), mean absolute percentage error (MAPE), and coefficient of determination ($R^2$). These metrics are reported separately for prefill latency, decode latency, and total latency. For transfer experiments, we report metrics before and after calibration to quantify the benefit of lightweight target-domain adaptation.
For deployment-oriented screening, the final model selection is based on the trade-off between predicted latency and model accuracy or task quality. Given a set of candidate models, the system combines the predicted latency $\hat{L}$ with an accuracy or quality score $A$ and identifies Pareto-optimal candidates. A model is retained if no other candidate achieves both lower predicted latency and higher accuracy. We report the number of candidate models, the number of retained models after screening, filtering ratio, Top-1 hit, Top-3 hit, feasible recall, and whether the best feasible model is preserved. This protocol evaluates the framework not only as a latency predictor, but also as a deployment-oriented model selection system.

\section{Results and Analysis}
\label{sec:results}

We evaluate the proposed framework from three perspectives: latency prediction accuracy, transferability under calibration, and deployment-oriented model screening. In addition, we validate whether the request-level profiling schema can be extended beyond mobile devices to heterogeneous edge and GPU platforms. This organization mirrors the system design: source-device profiling collects device profiles, model descriptors, deployment configurations, and runtime traces; the predictor estimates prefill, decode, and total latency; calibration adapts the predictor to target-device runtime behavior; and deployment screening uses predicted latency, together with model quality, to identify promising candidate models.

\subsection{Dataset and Profiling Coverage}

Table~\ref{tab:profiling_coverage} summarizes the dataset coverage used in the evaluation. The Pixel mobile dataset provides the main quantitative prediction results, including within-device prediction, cross-device transfer, calibration, ablation, and deployment screening. The Jetson Nano, Orange Pi 5 Pro, and RTX~3090-class datasets are used to validate heterogeneous profiling coverage. After excluding Qwen-family models from the reported model pool, the evaluation includes 515 request-level records across four device classes.
The dataset supports two levels of evaluation. First, the Pixel mobile dataset supports quantitative predictor evaluation because it includes cleaned request-level samples, phase-level latency targets, and complete evaluation results. Second, the Jetson Nano, Orange Pi 5 Pro, and RTX~3090-class datasets demonstrate that the profiling pipeline can collect compatible request-level records across different runtime and hardware environments. These non-mobile datasets are therefore used to validate the extensibility of the profiling schema rather than to claim final cross-device predictor-training results across all platforms.

\begin{table*}[t]
\centering
\caption{Dataset and profiling coverage used in the evaluation.}
\label{tab:profiling_coverage}
\footnotesize
\setlength{\tabcolsep}{5pt}
\renewcommand{\arraystretch}{1.12}
\begin{tabular}{p{0.22\textwidth} c c c p{0.36\textwidth}}
\toprule
Device & Requests & Models & Prompts & Use in evaluation \\
\midrule
Pixel~8 / Pixel~8~Pro 
& 260 & 5 & 43 
& Main prediction dataset for within-device prediction, mobile cross-device transfer, calibration, ablation, and deployment screening \\
Jetson Nano 
& 55 & 4 & 5 
& Heterogeneous profiling validation on an embedded GPU edge platform \\
Orange Pi 5 Pro 
& 155 & 1 & 155 
& Heterogeneous profiling validation on a single-board edge platform using an MLC-compatible SmolLM2 model \\
RTX~3090-class GPU 
& 45 & 5 & 3 
& Heterogeneous profiling validation on a high-performance GPU reference platform \\
\bottomrule
\end{tabular}
\end{table*}

\subsection{Within-Mobile-Device Latency Prediction}

We first evaluate the predictor in a within-device setting, where training and testing are performed on the same mobile device. Table~\ref{tab:within_device_results} compares the static-only baseline and the full predictor for prefill, decode, and total latency.

\begin{table*}[t]
\centering
\caption{Within-device latency prediction results on Pixel mobile devices. Lower is better for MAE/RMSE/MAPE, and higher is better for $R^2$.}
\label{tab:within_device_results}
\footnotesize
\setlength{\tabcolsep}{4pt}
\renewcommand{\arraystretch}{1.08}
{
\begin{tabular}{lllccccc}
\toprule
Device & Target & Method & MAE (ms) & RMSE (ms) & MAPE (\%) & $R^2$ & Train/Test \\
\midrule
Pixel 8 & Prefill & Static-only & 168.45 & 267.69 & 227.56 & 0.489 & 45 / 11 \\
Pixel 8 & Prefill & Full model & \textbf{132.24} & \textbf{256.00} & \textbf{35.57} & \textbf{0.532} & 45 / 11 \\
Pixel 8 & Decode & Static-only & 785.40 & 991.52 & 12.20 & 0.957 & 45 / 11 \\
Pixel 8 & Decode & Full model & \textbf{678.06} & \textbf{787.29} & \textbf{10.76} & \textbf{0.973} & 45 / 11 \\
Pixel 8 & Total & Static-only & 850.80 & 1088.06 & 12.50 & 0.953 & 45 / 11 \\
Pixel 8 & Total & Full model & \textbf{819.36} & \textbf{995.38} & \textbf{11.87} & \textbf{0.960} & 45 / 11 \\
\midrule
Pixel 8 Pro & Prefill & Static-only & 79.35 & 185.85 & 52.84 & -1.383 & 163 / 41 \\
Pixel 8 Pro & Prefill & Full model & \textbf{7.03} & \textbf{22.08} & \textbf{2.33} & \textbf{0.966} & 163 / 41 \\
Pixel 8 Pro & Decode & Static-only & 94.81 & 138.70 & 5.38 & 0.981 & 163 / 41 \\
Pixel 8 Pro & Decode & Full model & \textbf{86.47} & \textbf{120.98} & \textbf{4.62} & \textbf{0.985} & 163 / 41 \\
Pixel 8 Pro & Total & Static-only & \textbf{109.65} & \textbf{146.82} & \textbf{6.06} & \textbf{0.978} & 163 / 41 \\
Pixel 8 Pro & Total & Full model & 128.39 & 179.53 & 6.12 & 0.967 & 163 / 41 \\
\bottomrule
\end{tabular}
}
\end{table*}

The within-device results show that the full predictor is effective for mobile latency estimation. On Pixel~8, the full model improves all three latency targets over the static-only baseline. Decode MAE decreases from 785.40\,ms to 678.06\,ms, and total latency MAE decreases from 850.80\,ms to 819.36\,ms. The corresponding $R^2$ values also improve from 0.957 to 0.973 for decode and from 0.953 to 0.960 for total latency.
On Pixel~8~Pro, the strongest improvement appears in prefill latency. The full model reduces prefill MAE from 79.35\,ms to 7.03\,ms and improves $R^2$ from -1.383 to 0.966. This result shows the value of runtime-aware and phase-aware features for short and bursty execution phases. Decode latency also improves, although the gain is smaller because the static-only baseline is already strong on this device. For total latency on Pixel~8~Pro, the static-only baseline slightly outperforms the full model, suggesting that runtime-aware features may introduce noise when static descriptors already explain most of the latency variation.
Overall, decode and total latency are easier to predict than prefill latency. This is consistent with the execution structure of LLM inference: decode contains repeated autoregressive steps, while prefill is shorter, burstier, and more sensitive to trace alignment. The stronger Pixel~8~Pro result is also influenced by data coverage: the Pixel~8~Pro split uses 163 training samples, while Pixel~8 uses only 45.

\subsection{Mobile Cross-Device Transfer and Calibration}

We next evaluate cross-device transfer between Pixel~8 and Pixel~8~Pro. Table~\ref{tab:cross_device_transfer} reports zero-shot transfer results before calibration and calibrated results after using a small target-device calibration set.

\begin{table*}[t]
\centering
\caption{Cross-device transfer results before and after calibration.}
\label{tab:cross_device_transfer}
\footnotesize
\setlength{\tabcolsep}{4pt}
\renewcommand{\arraystretch}{1.08}
{
\begin{tabular}{lllccccc}
\toprule
Source & Target Device & Target / Calibration & MAE (ms) & RMSE (ms) & MAPE (\%) & $R^2$ & Train / Cal / Test \\
\midrule
Pixel 8 & Pixel 8 Pro & Prefill / Before & 101.24 & 205.35 & 51.73 & 0.374 & 56 / 20 / 184 \\
Pixel 8 & Pixel 8 Pro & Prefill / After  & 103.15 & 193.43 & 56.45 & 0.444 & 56 / 20 / 184 \\
Pixel 8 & Pixel 8 Pro & Decode / Before  & 6884.49 & 7634.37 & 443.61 & -33.237 & 56 / 20 / 184 \\
Pixel 8 & Pixel 8 Pro & Decode / After  & \textbf{575.53} & \textbf{980.45} & \textbf{32.18} & \textbf{0.435} & 56 / 20 / 184 \\
Pixel 8 & Pixel 8 Pro & Total / Before   & 6848.20 & 7640.54 & 390.72 & -28.232 & 56 / 20 / 184 \\
Pixel 8 & Pixel 8 Pro & Total / After   & \textbf{664.37} & \textbf{1087.11} & \textbf{32.38} & \textbf{0.408} & 56 / 20 / 184 \\
\midrule
Pixel 8 Pro & Pixel 8 & Prefill / Before & 50.12 & 77.49 & 162.13 & 0.769 & 204 / 6 / 50 \\
Pixel 8 Pro & Pixel 8 & Prefill / After  & \textbf{19.09} & \textbf{54.82} & \textbf{23.92} & \textbf{0.884} & 204 / 6 / 50 \\
Pixel 8 Pro & Pixel 8 & Decode / Before  & 5912.95 & 6994.85 & 76.38 & -1.085 & 204 / 6 / 50 \\
Pixel 8 Pro & Pixel 8 & Decode / After  & \textbf{1013.59} & \textbf{1307.18} & \textbf{20.99} & \textbf{0.927} & 204 / 6 / 50 \\
Pixel 8 Pro & Pixel 8 & Total / Before   & 5757.56 & 6864.12 & 72.87 & -0.974 & 204 / 6 / 50 \\
Pixel 8 Pro & Pixel 8 & Total / After   & \textbf{992.04} & \textbf{1200.06} & \textbf{18.23} & \textbf{0.940} & 204 / 6 / 50 \\
\bottomrule
\end{tabular}
}
\end{table*}

The transfer results show that calibration is essential. Without calibration, decode and total latency transfer poorly, with strongly negative $R^2$ values in several cases. After calibration, the predictor becomes usable across all three latency targets. For Pixel~8 $\rightarrow$ Pixel~8~Pro, calibration improves decode latency from $R^2=-33.237$ to $R^2=0.435$ and total latency from $R^2=-28.232$ to $R^2=0.408$. For Pixel~8~Pro $\rightarrow$ Pixel~8, calibration improves decode latency from $R^2=-1.085$ to $R^2=0.927$ and total latency from $R^2=-0.974$ to $R^2=0.940$.
Transfer is also asymmetric. Pixel~8~Pro $\rightarrow$ Pixel~8 performs much better than Pixel~8 $\rightarrow$ Pixel~8~Pro after calibration. This is consistent with the difference in source-device coverage: the Pixel~8~Pro source setting uses 204 training samples, whereas the Pixel~8 source setting uses only 56. These results align with the system design: the source-trained predictor provides a reusable latency prior, but the calibration set is needed to anchor the predictor to the target-device runtime regime.

\subsection{Heterogeneous Device Profiling Validation}

To validate the source-device profiling stage across different runtime environments, we analyze request-level profiling results from Jetson Nano, Orange Pi 5 Pro, and the RTX~3090-class GPU platform. These experiments are not used as final predictor-training results. Instead, they verify that the profiling pipeline can collect phase-level latency and runtime telemetry across heterogeneous devices while excluding Qwen-family models from the reported model pool.

\begin{table*}[t]
\centering
\caption{Heterogeneous profiling validation across non-mobile platforms. Latency and throughput values are averaged over request-level samples.}
\label{tab:heterogeneous_profiling}
\footnotesize
\setlength{\tabcolsep}{4pt}
\renewcommand{\arraystretch}{1.08}
{
\begin{tabular}{llrrrrr}
\toprule
Device & Model & Requests & Prompts & Avg. prefill (ms) & Avg. total latency (ms) & Decode throughput (tok/s) \\
\midrule
Jetson Nano & Tiny-GPT2 local & 20 & 5 & 7.15 & 86.90 & 201.68 \\
Jetson Nano & DistilGPT2 & 15 & 5 & 469.77 & 4364.14 & 4.11 \\
Jetson Nano & GPT-2 & 15 & 5 & 736.97 & 6694.74 & 2.69 \\
Jetson Nano & GPT-Neo-125M local & 5 & 5 & 834.15 & 8194.22 & 2.17 \\
\midrule
Orange Pi 5 Pro & SmolLM2-1.7B-Instruct-q4f16\_1-MLC & 155 & 155 & 4519.50 & 10348.87 & 8.42 \\
\midrule
RTX~3090-class GPU & OPT-350M & 9 & 3 & 26.20 & 689.63 & 95.36 \\
RTX~3090-class GPU & SmolLM2-1.7B-Instruct & 9 & 3 & 33.96 & 1013.64 & 64.38 \\
RTX~3090-class GPU & TinyLlama-1.1B-Chat & 9 & 3 & 31.18 & 1046.30 & 62.39 \\
RTX~3090-class GPU & SmolLM2-135M-Instruct & 9 & 3 & 31.74 & 1207.88 & 53.65 \\
RTX~3090-class GPU & SmolLM2-360M-Instruct & 9 & 3 & 45.18 & 1422.06 & 45.96 \\
\bottomrule
\end{tabular}
}
\end{table*}

Table~\ref{tab:heterogeneous_profiling} shows that latency behavior varies substantially across device-runtime-model configurations. On Jetson Nano, Tiny-GPT2 is much faster than the other models, achieving 86.90\,ms average total latency and 201.68 tokens/s, while GPT-Neo-125M reaches 8194.22\,ms total latency and 2.17 tokens/s. This large spread shows that model identity and runtime implementation strongly affect edge-device latency.
The Orange Pi 5 Pro data validates the profiling pipeline on a single-board edge platform using an MLC-compatible SmolLM2 model. Compared with the RTX~3090-class GPU platform, the same SmolLM2 model family exhibits substantially higher latency on Orange Pi 5 Pro, with an average total latency of 10348.87\,ms and a decode throughput of 8.42 tokens/s. This gap reflects the large difference between a high-performance desktop GPU and a lightweight edge board, and it motivates the need for device-aware latency prediction rather than relying on model descriptors alone.
The RTX~3090 results validate the high-performance GPU profiling path. The pipeline records phase-level latency, GPU utilization, power, temperature, clocks, memory usage, and estimated GPU-side energy. A notable observation is that smaller models are not always faster on the RTX~3090-class GPU. For example, SmolLM2-135M and SmolLM2-360M are slower than SmolLM2-1.7B in this run. This indicates that GPU utilization, kernel launch overhead, memory behavior, and runtime implementation can dominate simple parameter-count scaling. These observations support the need for both static and dynamic predictor paths.

\subsection{Ablation Study}

To understand the contribution of the main predictor components, Table~\ref{tab:ablation_results} compares four variants on the Pixel~8 $\rightarrow$ Pixel~8~Pro transfer direction: a static-only predictor, a version without phase-aware prefill features, the full model without calibration, and the full model with calibration.

\begin{table*}[t]
\centering
\caption{Ablation results on Pixel~8 $\rightarrow$ Pixel~8~Pro transfer.}
\label{tab:ablation_results}
\footnotesize
\setlength{\tabcolsep}{4pt}
\renewcommand{\arraystretch}{1.08}
{
\begin{tabular}{llrrrr}
\toprule
Target & Variant & MAE (ms) & RMSE (ms) & MAPE (\%) & $R^2$ \\
\midrule
Prefill & Static-only & 180.71 & 317.29 & 103.22 & -0.496 \\
Prefill & No phase-aware prefill features & 116.98 & 226.14 & 59.56 & 0.240 \\
Prefill & Full w/o calibration & 101.24 & 205.35 & 51.73 & 0.374 \\
Prefill & Full w/ calibration & \textbf{103.15} & \textbf{193.43} & \textbf{56.45} & \textbf{0.444} \\
\midrule
Decode & Static-only & 6769.44 & 7761.78 & 374.83 & -34.390 \\
Decode & No phase-aware prefill features & 6805.68 & 7530.68 & 439.32 & -32.314 \\
Decode & Full w/o calibration & 6884.49 & 7634.37 & 443.61 & -33.237 \\
Decode & Full w/ calibration & \textbf{575.53} & \textbf{980.45} & \textbf{32.18} & \textbf{0.435} \\
\midrule
Total & Static-only & 6674.27 & 7680.50 & 331.29 & -28.538 \\
Total & No phase-aware prefill features & 6789.94 & 7589.14 & 385.30 & -27.840 \\
Total & Full w/o calibration & 6848.20 & 7640.54 & 390.72 & -28.232 \\
Total & Full w/ calibration & \textbf{664.37} & \textbf{1087.11} & \textbf{32.38} & \textbf{0.408} \\
\bottomrule
\end{tabular}
}
\end{table*}

The ablation results support two conclusions. First, static descriptors alone are not sufficient for robust cross-device transfer. The static-only predictor performs poorly across all targets in the Pixel~8 $\rightarrow$ Pixel~8~Pro setting, especially for decode and total latency. Second, calibration is the most important factor for stabilizing transfer under the current data regime. Even the full model without calibration fails for decode and total latency, while the calibrated full model reduces total latency MAE from 6848.20\,ms to 664.37\,ms and improves $R^2$ from -28.232 to 0.408.
The phase-aware features provide clear benefit for prefill latency, improving $R^2$ from -0.496 for the static baseline to 0.240 without calibration and 0.374 in the full uncalibrated model. However, phase-aware modeling alone cannot solve cross-device mismatch. This confirms the role of the calibration set in the system design: static and dynamic representations provide a transferable prior, while calibration aligns the predictor with the target-device latency scale.

\subsection{Deployment-Oriented Screening}

Finally, we evaluate whether predicted latency can support deployment-time model screening. In this setting, the system uses predicted total latency to filter candidate models before real target-device execution. In the full system, the final model selection is based on the Pareto trade-off between predicted latency and model accuracy or task quality. The screening results focus on the latency side of this decision process and evaluate whether the predictor preserves the strongest feasible candidates.

\begin{table*}[t]
\centering
\caption{Combined screening results for total latency.}
\label{tab:screening_combined}
\footnotesize
\setlength{\tabcolsep}{6pt}
\renewcommand{\arraystretch}{1.12}
\begin{tabular}{lccccc}
\toprule
Source $\rightarrow$ Target & Retained / Total & Filter Ratio & Top-1 Correct & Top-3 Hit & Best Feasible Preserved \\
\midrule
Pixel 8 $\rightarrow$ Pixel 8 Pro & 2 / 3 & 0.333 & 0 & 1 & 1 \\
Pixel 8 Pro $\rightarrow$ Pixel 8 & 1 / 2 & 0.500 & 1 & 1 & 1 \\
\bottomrule
\end{tabular}
\end{table*}

Table~\ref{tab:screening_combined} shows that the predictor can reduce the candidate set while preserving the best feasible model. In the Pixel~8 $\rightarrow$ Pixel~8~Pro direction, the system retains 2 out of 3 candidates, corresponding to a filtering ratio of 0.333. In the Pixel~8~Pro $\rightarrow$ Pixel~8 direction, it retains 1 out of 2 candidates, corresponding to a filtering ratio of 0.500. In both cases, the best feasible model is preserved.
The ranking quality is mixed. Pixel~8~Pro $\rightarrow$ Pixel~8 correctly identifies the top-1 model, while Pixel~8 $\rightarrow$ Pixel~8~Pro does not. However, the true strong candidate remains in the retained set, as indicated by the Top-3 hit and best-feasible-preserved results. This suggests that the predictor is useful as a pre-screening tool even when the exact final ranking is imperfect. Once model accuracy or task-quality scores are incorporated, this latency-based filtering stage can be extended into the full Pareto-based deployment selection described in the system design.

\subsection{Discussion}

The results support four main observations. First, within-device latency prediction is strong on mobile devices, especially for decode and total latency. Second, cross-device transfer is feasible but requires calibration; without a small target-device calibration set, decode and total latency can shift too much across devices. Third, the ablation study confirms that static descriptors alone are insufficient and that calibration is essential for practical transfer. Fourth, the heterogeneous profiling results show that the request-level data schema can be extended beyond mobile devices to embedded GPU, single-board edge, and desktop GPU platforms.

The heterogeneous profiling results also highlight why the predictor needs both static and dynamic paths. Model size alone does not reliably determine latency. Smaller models can be slower than larger ones on high-performance GPUs, and the same model family can behave very differently across device classes. These effects arise from runtime backend, kernel efficiency, utilization, memory behavior, prompt characteristics, and generation length. Therefore, latency prediction must model the full device-runtime-model-prompt configuration.
At the same time, the evaluation has limitations. The most complete prediction results are based on the Pixel mobile dataset, while Jetson Nano, Orange Pi 5 Pro, and RTX~3090 are used mainly for profiling validation. The candidate pool for deployment screening is also small, and the screening table focuses on latency rather than a complete accuracy-latency Pareto frontier. Future experiments will combine all profiled devices into a unified training pipeline, increase prompt and model coverage on each platform, and evaluate full cross-device, cross-model, and cross-prompt splits with Pareto-based latency-quality model selection.
\section{Related Work}

\subsection{On-Device and Edge LLM Deployment}

Recent advances in runtime optimization and system support have made it increasingly feasible to deploy large language models on mobile and edge platforms~\cite{li2024mobilellm,xiao2024pockets,xu2024ondevice,mlcllm2024github}. However, practical deployment still requires deciding which model is suitable for which device under which workload constraints. Our work addresses this problem from the perspective of latency-aware model screening rather than runtime design alone.

\subsection{Latency Prediction and Transferability}

Latency prediction has long been used to reduce the cost of hardware-aware model deployment. Prior work such as nn-Meter demonstrates the value of predictive latency modeling over exhaustive benchmarking~\cite{zhang2021nnmeter}, while later studies show that static predictors are often limited by platform-specific behavior and runtime effects~\cite{akhauri2024latency}. More recent work emphasizes transferability, showing that latency or energy predictors should adapt to unseen hardware with limited additional profiling~\cite{feng2024litepred,lee2021help,tu2025platformx}. Our work builds on this direction, but focuses specifically on pretrained LLM deployment on heterogeneous edge devices.

\subsection{Runtime-Aware and Phase-Aware Performance Modeling}

For LLM inference, latency is shaped not only by static model and hardware properties, but also by execution structure and runtime state. Prior systems have shown that prefill and decode exhibit substantially different behaviors and should often be treated separately~\cite{yu2022orca,zhong2024distserve,kwon2023pagedattention}. At the hardware level, edge-device performance is also affected by DVFS, thermal throttling, and workload interference~\cite{kim2021ztt,lin2023geardvfs,benoitcattin2020thermal}. Our framework combines these two perspectives by integrating phase-aware trace modeling with runtime-aware latency prediction. For LLM inference, latency is shaped not only by static model and hardware properties, but also by execution structure and runtime state. Recent on-device LLM profiling work further shows that fine-grained runtime latency measurement can expose phase- and kernel-level bottlenecks on mobile devices~\cite{wang2025lmmeter}.

\subsection{Latency-Aware Model Screening}

Our work is also related to hardware-aware model selection. Prior benchmark efforts show that realistic hardware behavior cannot be captured by architecture descriptors alone~\cite{li2021hwnasbench}. Prior automated edge-AI design systems, such as GreenAuto and PlatformX, use profiling, prediction, and Pareto-based search to reduce the cost of sustainable model design on edge devices~\cite{tu2025greenauto,tu2025platformx}. Unlike these systems, which focus primarily on energy-efficient DNN/NAS design, we target screening pretrained LLM candidates from model zoos under device-specific and task-specific latency constraints. In this sense, we treat latency prediction as a practical system capability for deployment-time model selection.

\section{Conclusion}

We presented a runtime-aware and transferable framework for latency-aware LLM model screening on heterogeneous edge devices. By combining structured deployment descriptors, on-device profiling, phase-aware runtime modeling, and target-device calibration, the framework predicts prefill, decode, and total latency for candidate deployments.
Our results show that latency prediction for on-device LLM deployment must be both runtime-aware and transferable. Runtime-aware modeling improves the representation of real execution behavior, while lightweight calibration makes cross-device deployment more practical without exhaustive profiling of every candidate model on every platform.
More broadly, this work frames latency prediction as part of a deployment loop rather than as an isolated regression problem. We believe this perspective is important for scalable on-device LLM deployment, especially as model zoos continue to grow and edge platforms become increasingly diverse. Future work will extend the framework beyond latency to jointly consider other deployment objectives such as throughput, memory, and energy.

\section{Acknowledgement}

Research was sponsored by the Army Research Laboratory and was accomplished under Cooperative Agreement Number W911NF-23-2-0224. The views and conclusions contained in this document are those of the authors and should not be interpreted as representing the official policies, either expressed or implied, of the Army Research Laboratory or the U.S. Government. The U.S. Government is authorized to reproduce and distribute reprints for Government purposes notwithstanding any copyright notation herein.

\bibliographystyle{unsrt}
\bibliography{references}

\end{document}